\documentclass{article}

\usepackage{Arxiv}

\usepackage[utf8]{inputenc} 
\usepackage[T1]{fontenc}    
\usepackage{hyperref}       
\usepackage{url}            
\usepackage{booktabs}       
\usepackage{amsfonts}       
\usepackage{nicefrac}       
\usepackage{microtype}      
\usepackage{lipsum}
\usepackage{fancyhdr}       
\usepackage{graphicx}       
\graphicspath{{media/}}     

\title{\textbf{An AI-Driven Live Systematic Reviews in the Brain-Heart Interconnectome: \\
Minimizing Research Waste and Advancing Evidence Synthesis}}

\author{
    \textbf{A. Rahgozar}\textsuperscript{1}, 
    \textbf{P. Mortezaagha}\textsuperscript{1}, 
    \textbf{J. Edwards}\textsuperscript{1}, 
    \textbf{D. Manuel}\textsuperscript{1}, 
    \textbf{J. McGowen}\textsuperscript{1}, 
    \textbf{M. Zwarenstein}\textsuperscript{1}, 
    \textbf{D. Fergusson}\textsuperscript{1}, \\
    \textbf{A. Tricco}\textsuperscript{1}, 
    \textbf{K. Cobey}\textsuperscript{1}, 
    \textbf{M. Sampson}\textsuperscript{1}, 
    \textbf{M. King}\textsuperscript{1}, 
    \textbf{D. Richards}\textsuperscript{1}, 
    \textbf{A. Bodnaruc}\textsuperscript{1}, 
    \textbf{D. Moher}\textsuperscript{1} \\
    \\
    \textsuperscript{1}{Ottawa Hospital Research Institute, Ottawa, Ontario, Canada}
}

\begin{document}
\maketitle

\begin{abstract}
\noindent
\textbf{Background:} 
The rapidly evolving field of the Brain-Heart Interconnectome (BHI) merges neurology and cardiology. Despite its potential, inefficiencies in evidence synthesis and suboptimal adherence to quality standards often result in research waste. Systematic reviews are prone to redundancy, incomplete reporting, and lack of methodological rigor. 

\noindent
\textbf{Objectives:}
We introduce an artificial intelligence (AI)-driven system designed to streamline systematic reviews in the BHI domain. The system applies a multipronged approach that includes automated Population, Intervention, Comparator, Outcome, Study design (PICOS) detection, semantic search with vector embeddings, graph-based querying of complex relationships, topic modeling for redundancy assessment, and a continuously updated ``living'' database. 

\noindent
\textbf{Methods:}
Key components of the system include a Bi-directional Long Short-Term Memory (Bi-LSTM) model to assess PICOS compliance, Retrieval-Augmented Generation (RAG) and large language models (LLMs) for enhanced semantic retrieval, and BERTopic for thematic clustering. Graph-structured data are stored using Neo4j to capture interactions among interventions, outcomes, and study designs. Performance was evaluated through comparative analyses of RAG with GPT-3.5 versus plain GPT-4, with additional attention to the Bi-LSTM model's accuracy for PICOS classification and a hierarchical study design classifier. 

\noindent
\textbf{Results:}
RAG combined with GPT-3.5 outperformed GPT-4 for specialized BHI queries requiring graph-based and topic-driven insights. The Bi-LSTM model achieved an 87\% accuracy in distinguishing PICOS-compliant articles, while the study design classifier attained a 95.7\% overall accuracy. By detecting research redundancies, spotlighting underexplored topics, and offering real-time updates, this system systematically reduces waste. Its user-oriented interface provides interactive dashboards and conversational AI, supporting dynamic evidence synthesis.

\noindent
\textbf{Conclusion:}
This AI-enhanced platform addresses well-known challenges in systematic reviews and research waste. Although originally built for BHI, its architectural flexibility—incorporating advanced NLP, machine learning, and graph-based analytics—renders it broadly adaptable to various biomedical research fields. By enabling timely and rigorous evidence synthesis, our approach facilitates more efficient research allocations, fosters high-quality methodological standards, and supports informed clinical decision-making.

\end{abstract}


\section{Introduction}
\label{sec:introduction}
The Brain-Heart Interconnectome (BHI) explores the intricate bidirectional relationships between the cardiovascular and neurological systems \cite{deking2021interactions, cai2024physiologic, sabor2022bhinet, catrambone2023complex}. While BHI holds considerable promise for improving patient care, inefficiencies in synthesizing evidence, coupled with inconsistent adherence to quality standards, create obstacles for both clinical and research endeavors. These inefficiencies often give rise to \textit{research waste}, defined as redundant or low-quality studies that fail to enhance collective knowledge \cite{chalmers2009avoidable, chapman2019discontinuation, alexander2020research, rosengaard2024methods}.

Research waste persists at virtually all stages of evidence synthesis. For instance, recent studies show that more than 85\% of surgical randomized controlled trials (RCTs) exhibit at least one form of research waste—ranging from inadequate reporting to non-publication \cite{chapman2019discontinuation}—and comparable shortcomings persist in other areas of biomedical research \cite{lu2021reporting}. Such inefficiencies generate a cycle in which researchers, clinicians, and policymakers often lack the most rigorous and current data necessary for optimal decisions \cite{dalsanto2022research, yordanov2018avoidable}.

Methodological frameworks like PICOS (Population, Intervention, Comparator, Outcome, Study design) \cite{wei2024pico, zhang2023picos, lei2023picos} promise improved coherence and replicability in clinical trial design \cite{zhang2023task, feldner2023exploring, dhrangadhariya2024pico}. However, evidence suggests that guidelines such as CONSORT (Consolidated Standards of Reporting Trials) \cite{david_moher2024consort} and PRISMA (Preferred Reporting Items for Systematic Reviews and Meta-Analyses) \cite{tricco2021prisma} remain inconsistently applied, undermining their full potential. In particular, many systematic review authors report substantial challenges in managing extensive literature searches and ensuring meticulous screening and data extraction \cite{santo2022research}.

The shortcomings in current processes have prompted increasing interest in artificial intelligence (AI) tools to enhance systematic reviews. Several initiatives integrate AI-driven natural language processing (NLP) methods to expedite literature search, screening, data extraction, and risk-of-bias evaluations \cite{OforiBoateng2024revnlp, jin2018pico}. Although recent successes—such as RobotReviewer and SciSpace \cite{robotreviewer2024tools, petri2023evidence, ge2024aisysrev}—indicate that AI can lighten the manual workload, domain-specific customization and dynamic updating features remain underdeveloped in many of these systems \cite{menezes2023gpt4medicalnotes}.

\subsection{Aim and Scope of the Proposed System}
This paper introduces an AI-driven system designed to systematically enhance evidence synthesis in the BHI domain. Our solution incorporates:
\begin{itemize}
    \item \textbf{Automated PICOS Compliance Detection:} Prioritizes studies aligned with rigorous methodological standards.
    \item \textbf{Semantic Search \& Graph-Based Queries:} Leverages Neo4j \cite{rafael2024neo4j} and pgVector to reveal meaningful relational structures among diverse data elements (e.g., outcomes, interventions).
    \item \textbf{Topic Modeling via BERTopic:} Identifies core themes, flags redundant clusters, and highlights underexplored areas \cite{grootendorst2022bertopic}.
    \item \textbf{Living Database Functionality:} Dynamically integrates new publications into existing systematic reviews, ensuring near real-time updates.
    \item \textbf{Interactive Dashboards and Conversational AI:} Facilitates broad accessibility, enabling clinicians, researchers, and other stakeholders to engage with and interpret data more efficiently.
\end{itemize}

\subsection{Overall Workflow in Systematic Reviews}
\begin{figure}[ht]
  \centering
  \includegraphics[scale=0.47]{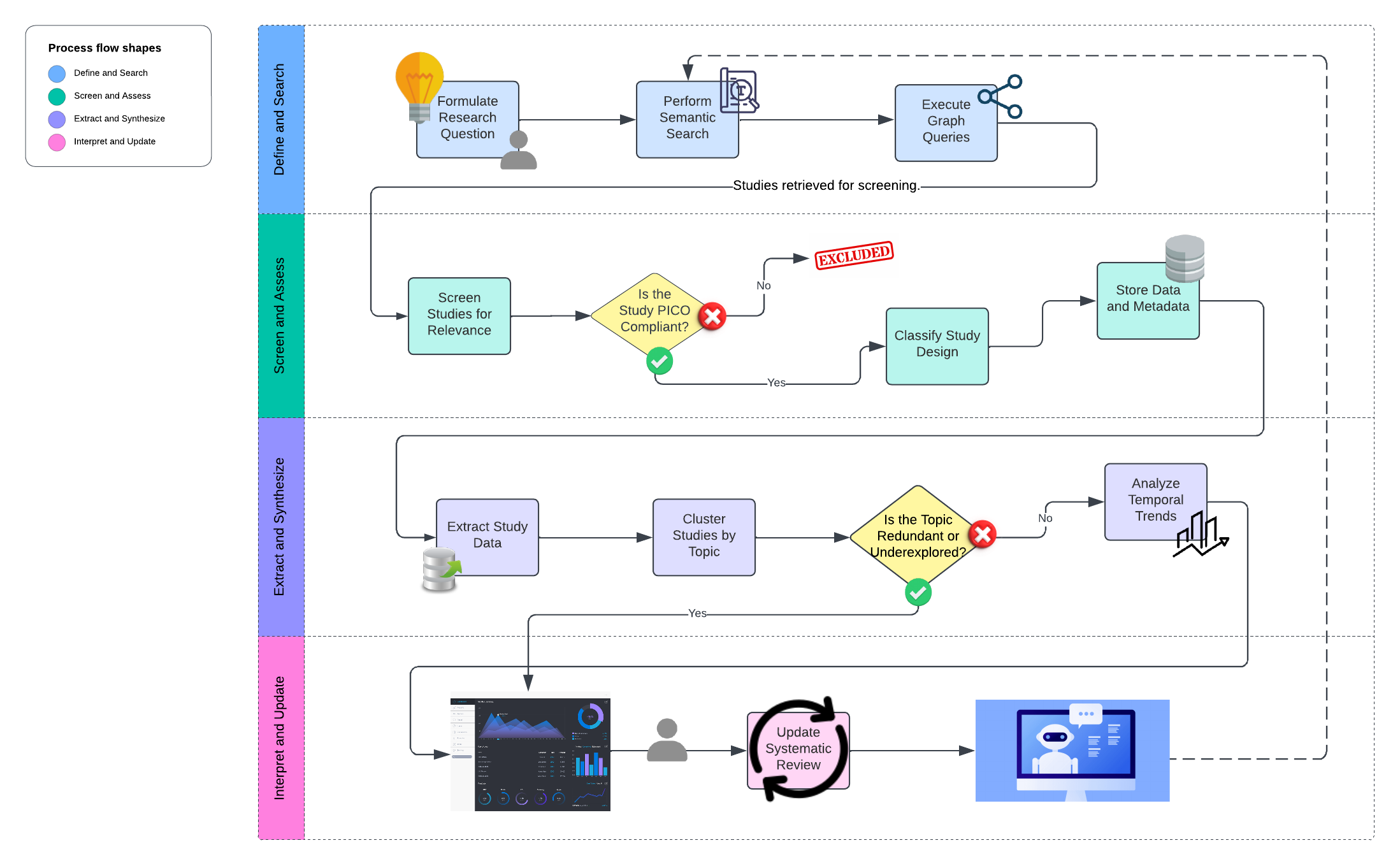}
  \caption{Workflow of the AI-Driven System for Systematic Reviews in BHI Research}
  \label{fig:sysrev_process_flow}
\end{figure}

Aligned with standard systematic review pipelines, the framework features four primary phases (Figure~\ref{fig:sysrev_process_flow}):
\begin{enumerate}
    \item \textit{Define and Search:} AI-guided query formulation based on PICOS, supported by semantic matching and graph-based retrieval.
    \item \textit{Screen and Assess:} Automated identification of PICOS-compliant articles and hierarchical classification of study designs (e.g., RCTs, cohort, case-control).
    \item \textit{Extract and Synthesize:} Dynamic clustering and topic modeling (using BERTopic) to highlight thematic areas, redundancy, and emerging trends.
    \item \textit{Interpret and Update:} Integration of new literature, real-time updates of the review database, and user interaction through dashboards and conversational AI.
\end{enumerate}

By minimizing redundancies and emphasizing high-value studies, this system aims to reduce time wasted on low-impact or duplicative research, thus improving the quality and efficiency of BHI-related evidence synthesis \cite{legate2024semiautomated, tomczyk2024ai, gorska2024towards, ofori2024towards, toth2024automation}.

\subsection{BHI Human Requirements}
\label{sec:bhi_human_requirements}

Beyond purely technical considerations, the system’s design and implementation were guided by iterative feedback from domain experts and key stakeholders in the Brain-Heart Interconnectome (BHI) community. Communication was primarily asynchronous (e.g., email exchanges), allowing for flexible collaboration across different time zones and schedules. These interactions served to:

\begin{itemize}
    \item \textbf{Clarify Screening Criteria:} Experts emphasized that studies lacking abstracts should not be excluded prematurely, recognizing that incomplete records can still hold value in emerging BHI areas. Eligibility criteria were further aligned with evidence-based frameworks to ensure consistency in participant selection, intervention design, and outcome evaluation. These criteria included:
    \begin{itemize}
        \item \textbf{Types of Participants:} Studies must involve human participants diagnosed with conditions relevant to the BHI domain, using accepted diagnostic criteria.
        \item \textbf{Types of Interventions:} Interventions must be planned, structured, and replicable, targeting specific outcomes related to neurological or cardiovascular health, with measurable fitness or health components.
        \item \textbf{Comparator Interventions:} The design must include appropriate controls, such as no intervention, usual care, or alternative validated interventions, to ensure the robustness of outcome assessments.
        \item \textbf{Outcome Measures:} Both primary and secondary outcomes must be evaluated using validated tools to capture key metrics such as cognitive function, activities of daily living, and health-related quality of life.
        \item \textbf{Timing of Assessments:} Outcome evaluations must include immediate post-intervention assessments as well as short-term and long-term follow-ups to provide a comprehensive view of intervention efficacy.
    \end{itemize}
    \item \textbf{Refine Evaluation Metrics and Terminology:} Internal discussions reinforced the equivalence of “sensitivity” and “recall,” while highlighting additional metrics such as Positive Predictive Value (PPV) and Negative Predictive Value (NPV). These exchanges also confirmed that “YES/NO” labeling for study type predictions aligns with “TRUE/FALSE” classification conventions in clinical research.
    \item \textbf{Expand Classification Dimensions:} Stakeholders identified a “Setting” category (e.g., community, locked facility) necessary for systematically capturing the context of BHI investigations. This requirement led to enhancements in our hierarchical classification scheme, such as incorporating “systematic review” as a distinct study design node.
\end{itemize}

By incorporating structured eligibility criteria alongside expert-driven feedback, the system ensures consistency in evaluating studies while maintaining methodological rigor and practical relevance. As BHI research continues to evolve, ongoing collaboration with domain experts will remain integral to refining classification schemes, updating evaluation metrics, and aligning the system with real-world clinical and research needs.

\section{Methodology}
\label{sec:methods}
We developed a multifaceted architecture (Figure~\ref{fig:workflow}) that integrates cutting-edge AI and NLP components. The system coordinates retrieval, analysis, and exploration functions tailored specifically to the Brain-Heart Interconnectome (BHI) domain. Key components include:

\begin{figure}[h]
    \centering
    \includegraphics[width=0.7\textwidth]{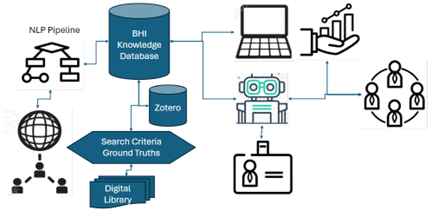}
    \caption{\centering Workflow for Knowledge Extraction and Querying}
    \label{fig:workflow}
\end{figure}

\subsection{Graph Database for Biomedical Insights}
\begin{figure}[ht]
    \centering
    \includegraphics[width=0.5\textwidth]{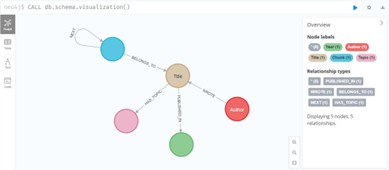}
    \caption{Neo4j Graph Visualization of Relationships}
    \label{fig:neo4j}
\end{figure}

Recognizing the inherent complexity of BHI-related data, we utilize Neo4j to represent relationships among key entities such as interventions, outcomes, author affiliations, and publication venues. These graph structures enable sophisticated Cypher queries that uncover nuanced patterns (e.g., comorbidities, multiple interventions within a single study), ensuring a more comprehensive understanding of the domain (Figure~\ref{fig:neo4j}) \cite{peng2024graphragsurvey}.

\subsection{Retrieval-Augmented Generation and Semantic Search}
To optimize document retrieval and streamline knowledge generation:
\begin{itemize}
    \item \textbf{pgVector (PostgreSQL Extension):} Stores document embeddings, enabling vector-based semantic searches that extend beyond simple keyword matching.
    \item \textbf{LangChain:} Coordinates query processing and retrieval, ensuring that system responses are contextually and thematically aligned with user input. 
\end{itemize}

We examined GPT-3.5 embedded within a RAG framework \cite{fan2024surveyragmeetingllms} and benchmarked it against standalone GPT-4. The RAG approach demonstrated superior performance in field-specific queries that require deeper data integration, such as comparing multiple interventions or focusing on specific outcome measures documented in the graph database.

\subsection{Automated PICO Compliance Detection and Study Design Classification}
\label{sec:pico_studydesign}
\begin{figure}[ht]
    \centering
    \includegraphics[width=0.38\textwidth]{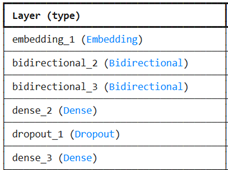}
    \caption{Architecture of the Bi-LSTM Model for PICO Compliance Detection}
    \label{fig:BiLSTM}
\end{figure}

\begin{enumerate}
    \item \textbf{PICO Compliance:} A Bi-LSTM model, trained on an extended PubMed-PICO dataset \cite{jin2018pico}, identifies whether a given abstract conforms to PICOS standards. The model consists of an embedding layer, followed by two bidirectional LSTM layers, two dense layers, and a dropout layer for regularization (Figure~\ref{fig:BiLSTM}). 
    \item \textbf{Study Design Classification:} Large language models (LLMs) are employed to determine study design hierarchies. This process yields precise categorization (e.g., RCT, cohort, case-control, systematic review), which is crucial for the rapid identification of high-quality, methodologically sound research. Classification outcomes are stored in PostgreSQL for future query-based filtering.
\end{enumerate}

\subsection{Topic Modeling with BERTopic}
BERTopic supports thematic clustering and identification of underexplored or redundant lines of inquiry. By storing topic labels in both PostgreSQL and Neo4j, the system facilitates flexible exploration and real-time visualization. Researchers can track how key research themes evolve and cluster around pivotal interventions or outcomes. The BERTopic “terms” that define each cluster also serve to highlight the principal concepts driving the conversation in each theme.

\subsubsection{Visualization and Dashboards}
\label{sec:viz}
\begin{figure}[ht]
    \centering
    \includegraphics[width=0.7\textwidth]{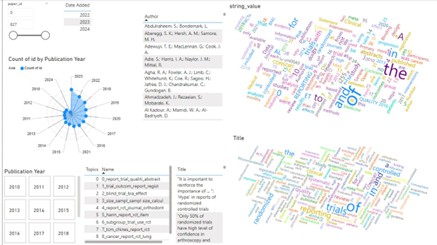}
    \caption{Power BI Dashboard Visualizing Research Trends}
    \label{fig:dashboard}
\end{figure}

\begin{figure}[ht]
    \centering
    \includegraphics[width=0.7\textwidth]{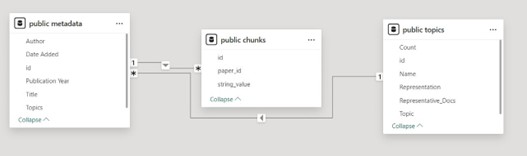}
    \caption{Relational Database Schema for Metadata, Document Chunks, and Topic Clusters}
    \label{fig:schema}
\end{figure}

A Power BI dashboard furnishes interactive visualizations (e.g., word clouds, radial charts) to elucidate publication patterns, authorship dynamics, and shifting research themes (Figures~\ref{fig:schema}, \ref{fig:dashboard}). In addition to supporting systematic reviewers, organizations such as the Canadian Institutes of Health Research (CIHR) can utilize this dashboard to monitor reporting guideline compliance (e.g., PICOS or CONSORT adherence) and to inform evidence-based policy and guidelines decisions. These functionalities allow for both granular and global views of BHI-related evidence, depending on user requirements.

\subsection{Modular Integration and Front-End Interface}
The system orchestrates data flow across multiple backends:
\begin{itemize}
    \item \textbf{LangGraph:} Directs complex, multi-step queries to the appropriate data store (graph, relational, or vector) while minimizing hallucinations by verifying document relevance.
    \item \textbf{LiteralAI API:} Logs all user interactions, promoting transparency and accountability.
    \item \textbf{Chainlit Frontend:} Offers a user-friendly, conversational interface. Domain experts and clinicians can pose natural language questions and receive curated responses supplemented by relevant source documents.
\end{itemize}

\subsection{Evaluation Criteria}
The system was assessed on:
\begin{enumerate}
    \item \textbf{Comparison: RAG vs.\ GPT-4.} Human experts rated the relevance and accuracy of system-generated answers. RAG-based GPT-3.5 outperformed plain GPT-4 in inquiries involving layered domain knowledge (e.g., synergy between interventions).
    \item \textbf{Quantitative Metrics:} Retrieval accuracy and content fidelity were measured using standard metrics (e.g., Mean Reciprocal Rank, ROUGE).
    \item \textbf{Hierarchical Study Design Classification:} Expert evaluations confirmed that the automated classification process was robust and aligned well with gold-standard references.
\end{enumerate}

\section{Results}
\label{sec:results}

\subsection{Performance of RAG vs.\ GPT-4}
\label{sec:performance_rag}
Expert reviewers evaluated RAG plus GPT-3.5 against GPT-4 across a series of domain-specific questions. Each query was scored on:
\begin{itemize}
    \item \textbf{Relevance of Retrieved Documents}
    \item \textbf{Accuracy and Contextual Fit}
    \item \textbf{Sufficiency for Guiding Clinical or Research Decisions}
\end{itemize}

Overall, 75\% of RAG-augmented responses met or exceeded expert expectations, with:
\begin{itemize}
    \item 30\% of queries adequately answered by both RAG-based GPT-3.5 and GPT-4.
    \item 25\% where RAG-based GPT-3.5 excelled, especially for relationship-centric queries leveraging Neo4j.
    \item 20\% where GPT-4’s broader generative capacity performed better (e.g., large-scale thematic overviews).
    \item 25\% where the system required further optimization, mainly due to complexity in synthesizing multiple data layers (e.g., PICOS compliance, graph-based relationships, and topic clusters).
\end{itemize}

Notably, the BERTopic “terms” that emerged within these topic clusters also helped refine certain queries. By focusing on the high-impact tokens identified in each cluster, RAG-based GPT-3.5 more effectively retrieved contextually relevant studies.

\subsection{Topic Modeling Outcomes}
\label{sec:topic_modeling_outcomes}
BERTopic-derived clusters helped researchers pinpoint areas with high publication density and detect thematic overlaps. In addition to summary statistics and top terms, multiple visual representations were generated to capture the topic distribution and how it evolves over time.

\begin{figure}[ht]
    \centering
    \includegraphics[width=0.65\textwidth]{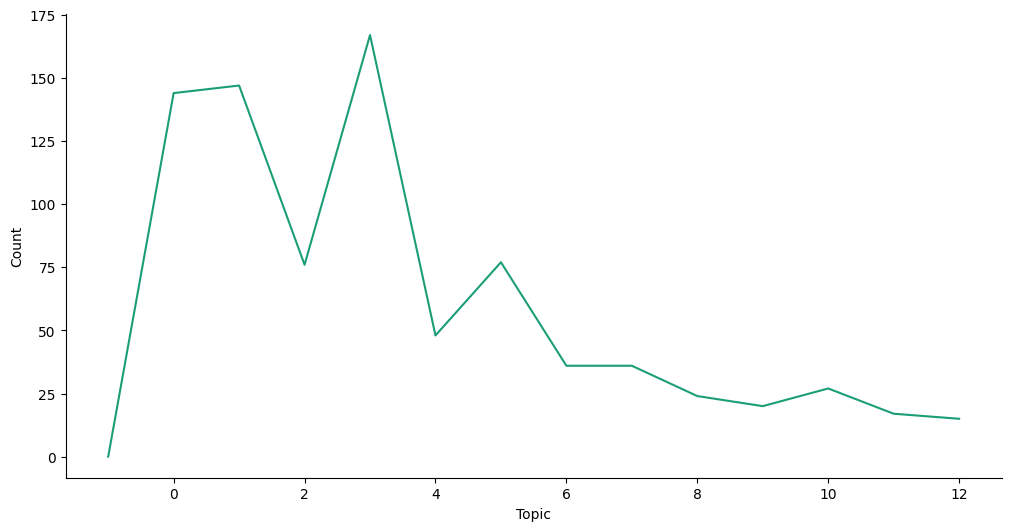}
    \caption{\centering Line chart showing the count of documents assigned to each discovered topic. Higher peaks indicate clusters with more publications.}
    \label{fig:topic_count}
\end{figure}

In Figure~\ref{fig:topic_count}, a line chart shows the number of documents assigned to each discovered topic. Peaks in this figure indicate clusters with a relatively higher number of publications. Table~\ref{tab:topic_summary} provides an illustrative overview of the topics, including the topic label, high-frequency terms, and an example of representative documents.

\begin{table}[ht]
\centering
\caption{\centering Illustrative BERTopic output for all discovered topics, showing label (Name), high-frequency terms (Representation), and count of documents.}
\label{tab:topic_summary}
\begin{tabular}{l l l l}
\toprule
\textbf{Topic} & \textbf{Count} & \textbf{Name} & \textbf{Representation} \\
\midrule

0 & -1 & \texttt{-1\_report\_trial\_studi\_rct} & 
\textit{\{report, trial, studi, rct, data, use, \dots\}} \\

1 & 0 & \texttt{0\_outcom\_trial\_regist\_registr} & 
\textit{\{outcom, trial, regist, registr, primari, \dots\}} \\

2 & 1 & \texttt{1\_blind\_trial\_bia\_effect} & 
\textit{\{blind, trial, bia, effect, outcom, \dots\}} \\

3 & 2 & \texttt{2\_harm\_report\_trial\_event} & 
\textit{\{harm, report, trial, event, advers, \dots\}} \\

4 & 3 & \texttt{3\_qualiti\_rct\_report\_abstract} & 
\textit{\{qualiti, rct, report, abstract, method, \dots\}} \\

5 & 4 & \texttt{4\_size\_sampl\_sampl size\_calcul} & 
\textit{\{size, sampl, sampl size, calcul, power, \dots\}} \\

6 & 5 & \texttt{5\_trial\_guidelin\_item\_report} & 
\textit{\{trial, guidelin, item, report, develop, \dots\}} \\

7 & 6 & \texttt{6\_journal\_report\_orthodont\_rct} & 
\textit{\{journal, report, orthodont, rct, publish, \dots\}} \\

8 & 7 & \texttt{7\_subgroup\_trial\_analys\_subgroup} & 
\textit{\{subgroup, trial, analys, subgroup analys, \dots\}} \\

9 & 8 & \texttt{8\_chines\_tcm\_report\_rct} & 
\textit{\{chines, tcm, report, rct, qualiti, \dots\}} \\

10 & 9 & \texttt{9\_pro\_rct\_cancer\_report} & 
\textit{\{pro, rct, cancer, report, qualiti, \dots\}} \\

11 & 10 & \texttt{10\_abstract\_report\_ci\_trial} & 
\textit{\{abstract, report, ci, trial, statist, \dots\}} \\

12 & 11 & \texttt{11\_recruit\_retent\_particip\_strategi} & 
\textit{\{recruit, retent, particip, strategi, \dots\}} \\

13 & 12 & \texttt{12\_trial\_particip\_patient\_organ} & 
\textit{\{trial, particip, patient, organ, platform, \dots\}} \\

\bottomrule
\end{tabular}
\end{table}

\begin{figure}[ht]
    \centering
    \includegraphics[width=0.65\textwidth]{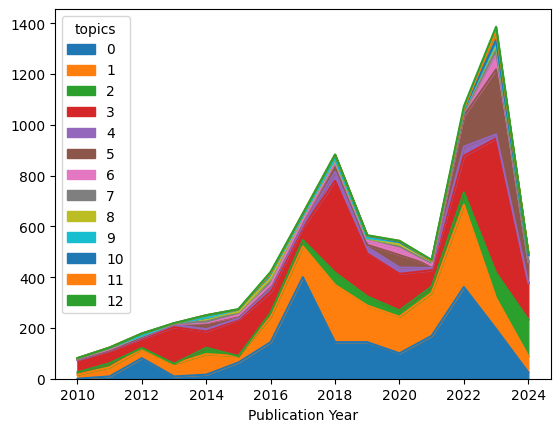}
    \caption{\centering Stacked area chart displaying how the topic composition shifts over time (2010--2024). Individual colors represent distinct topics.}
    \label{fig:topic_stacked_area}
\end{figure}

\begin{figure}[ht]
    \centering
    \includegraphics[width=0.9\textwidth]{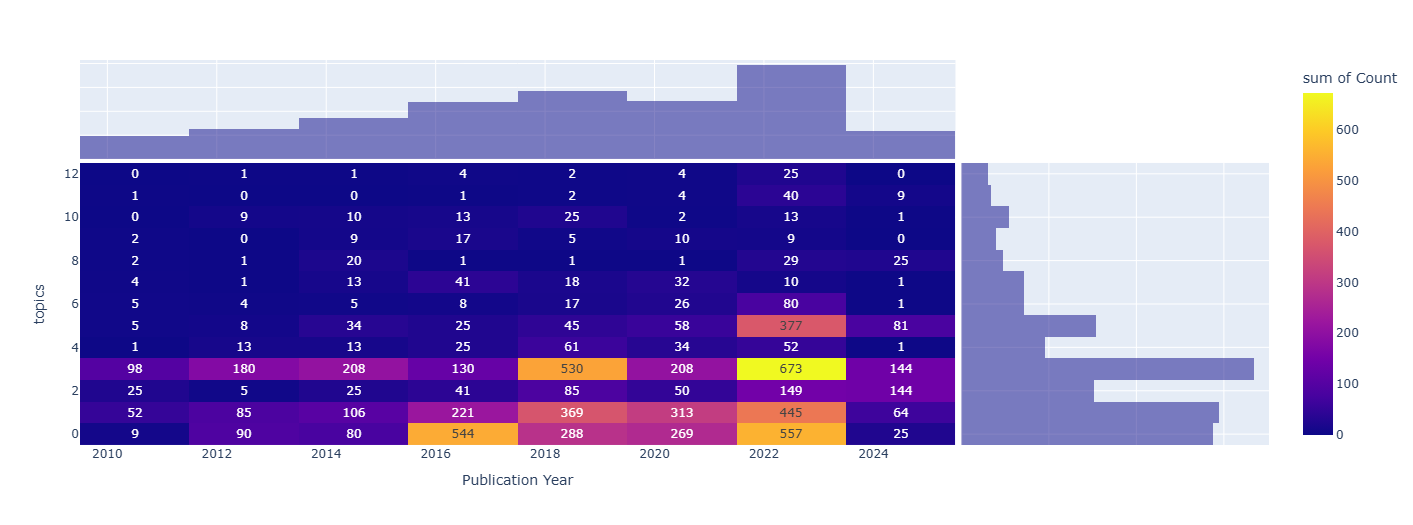}
    \caption{\centering Heatmap depicting the number of documents per topic (y-axis) per publication year (x-axis). Brighter (yellow) cells indicate higher document counts.}
    \label{fig:topic_heatmap}
\end{figure}

Figure~\ref{fig:topic_stacked_area} presents a stacked area chart depicting the temporal shifts in topic composition from 2010 to 2024. Each color represents a different topic, and changing proportions show how research interests evolve over time. Figure~\ref{fig:topic_heatmap} is a heatmap of document counts per topic (y-axis) by publication year (x-axis). Warmer (yellow) cells in this heatmap represent higher document frequencies.

\begin{figure}[ht]
    \centering
    \includegraphics[width=0.75\textwidth]{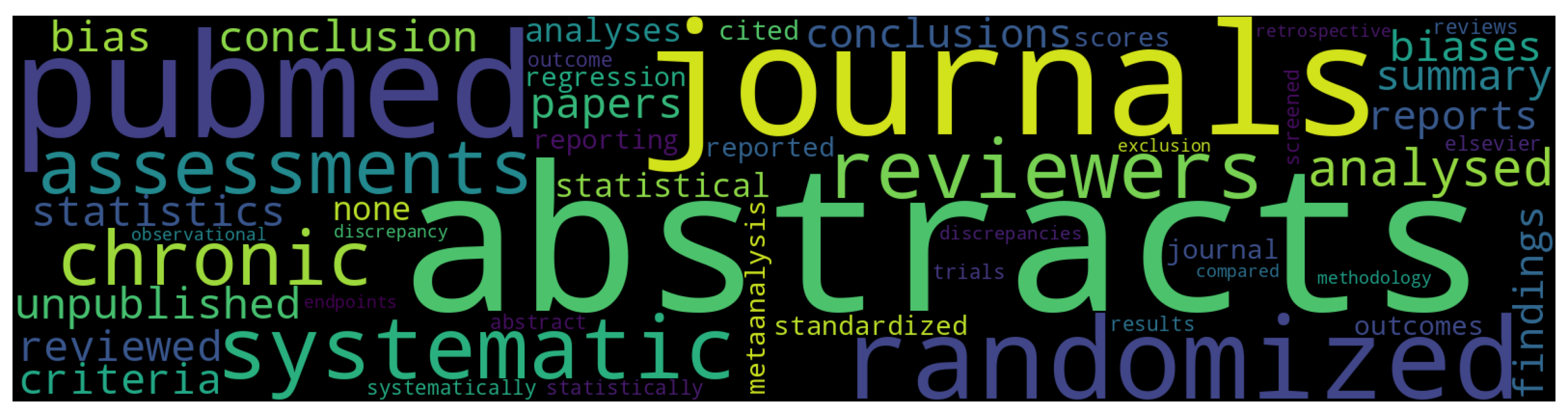}
    \caption{\centering Word cloud example}
    \label{fig:wordcloud_topic0}
\end{figure}

Figure~\ref{fig:wordcloud_topic0} shows an example of a word cloud, highlighting frequently occurring terms that point to methodological considerations such as “pubmed,” “journals,” and “randomized.” This provides a quick visual cue about the main ideas defining the cluster.

\textbf{Key Observations}
\begin{itemize}
    \item \textbf{Redundancy Alerts} were triggered by repeated short-term cardiovascular outcome studies, prompting a recommendation to investigate long-term neurocardiological impacts.  
    \item \textbf{Temporal Trends} can be visualized in Figures~\ref{fig:topic_stacked_area} and \ref{fig:topic_heatmap}, showing peaks in publication volume from 2018 to 2022. This surge may be tied to shifting funding priorities or increased adoption of reporting guidelines.  
    \item \textbf{Granularity in Analysis} is enhanced by combining BERTopic cluster terms with relational data, offering deeper insights into the interventions and outcomes dominating each topic.  
\end{itemize}

\subsection{Document Relevance and Hallucination Prevention}
\begin{figure}[ht]
    \centering
    \includegraphics[width=0.8\textwidth]{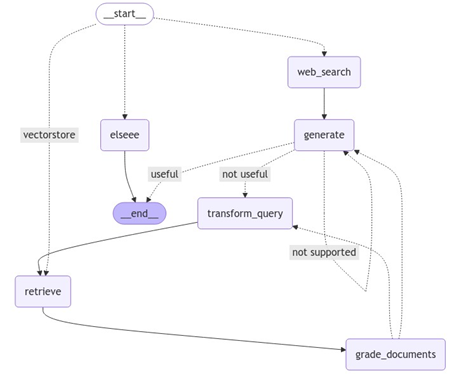}
    \caption{Workflow for Query Processing, Retrieval, and Relevance Grading}
    \label{fig:query_workflow}
\end{figure}

The LangGraph module proved successful at minimizing off-target references. By mapping retrieved texts back to original user queries, the system drastically reduced irrelevant or fabricated outputs commonly seen in standalone LLM-generated answers (Figure~\ref{fig:query_workflow}). This process enhances the clinical and research utility of AI-aggregated information.

\subsection{PICO Compliance Detection}
The Bi-LSTM model recorded an accuracy of 87\% in identifying PICOS-compliant abstracts. By spotlighting high-quality research early in the screening process, this feature helps curators avoid downstream inclusion of studies that fail to meet foundational evidence-based criteria.

\subsection{Study Design Classification}
A validation study of 164 references measured the system’s classification performance for study design. Table~\ref{tab:confusion_matrix} summarizes the confusion matrix.

\begin{table}[ht]
    \centering
    \renewcommand{\arraystretch}{1.2}
    \setlength{\tabcolsep}{12pt}
    \caption{Confusion Matrix for Study Design Prediction}
    \label{tab:confusion_matrix}
    \begin{tabular}{@{}lllll@{}}
        \toprule
        & True Positives & False Positives & True Negatives & False Negatives \\
        \midrule
        \textbf{Counts} & 74 & 7 & 83 & 0 \\
        \bottomrule
    \end{tabular}
\end{table}

Table~\ref{tab:performance_metrics} shows derived performance metrics.

\begin{table}[ht]
    \centering
    \renewcommand{\arraystretch}{1.2}
    \setlength{\tabcolsep}{12pt}
    \caption{Performance Metrics for Study Design Classification}
    \label{tab:performance_metrics}
    \begin{tabular}{@{}lllll@{}}
    \toprule
    \textbf{Metric} & \textbf{Precision} & \textbf{Recall} & \textbf{Specificity} & \textbf{Accuracy} \\
    \midrule
    Value & 91.4\% & 100\% & 92.2\% & 95.7\% \\
    \bottomrule
    \end{tabular}
\end{table}

These data underscore the system’s capability to rapidly identify and prioritize methodologically rigorous studies during the screening phase, augmenting the quality of subsequent analyses.

\section{Discussion}
\label{sec:discussion}
The Brain-Heart Interconnectome (BHI) promises significant breakthroughs in clinical practice and research. However, persistent issues of suboptimal data curation, incomplete reporting, and manual workflows that often fail to keep pace with new findings contribute to research waste \cite{alexander2020research, rosengaard2024methods}. Our proposed system seeks to alleviate these challenges by automating critical components of systematic reviews while providing decision support for researchers and clinicians.

\subsection{Addressing Redundancy and Guiding Research Focus}
\label{sec:addressing_redundancy}
By leveraging BERTopic to cluster related studies (see Section~\ref{sec:topic_modeling_outcomes}), the system exposes opportunities for methodological consolidation. The top terms extracted in each cluster help identify repetitive investigations. This feature aids editorial teams and funding bodies alike to redirect efforts toward genuinely novel or underexplored domains. In the BHI space, such direction could expedite the discovery of biomarkers, therapeutic targets, or improved patient screening methods.

\subsection{Quality Assurance Through PICO Compliance and Study Design}
Selective inclusion of PICOS-compliant and high-quality study designs (e.g., RCTs, systematic reviews for BHI) forms the foundation of dependable reviews. The Bi-LSTM model’s robust performance in screening abstracts for PICOS elements ensures that substandard studies are flagged early, thereby minimizing wasted effort in later review stages. Our hierarchical classification approach further refines selection by prioritizing the most robust methodologies within a given research theme.

\subsection{Dynamic Integration and Timely Evidence}
One major limitation of conventional reviews is the significant lag before newly published data are incorporated. This system’s living database structure updates continuously, ensuring that reviews reflect near real-time shifts in the evidence base. Such “live” systematic reviews can be of critical importance when time-sensitive decisions—such as those surrounding acute cardiovascular events or neuroprotective interventions—must be grounded in the latest evidence.

\subsection{Enhanced Accessibility via Conversational AI and Dashboards}
From an end-user perspective, the ability to pose domain-specific questions to a conversational AI dramatically lowers barriers to entry. Non-technical stakeholders—such as allied health professionals or policymakers—can navigate the evidence base without advanced data analytics expertise, thereby closing the gap between research production and application. The dynamic dashboard, enriched by BERTopic analyses, similarly empowers organizations to monitor compliance trends and uncover opportunities for targeted reporting guidelines and policy intervention.

\subsection{Potential for Broader Implementation}
Although optimized for BHI, the architecture is generalizable. Other specialties or cross-disciplinary fields encountering large, heterogeneous bodies of literature can benefit from this structured, AI-driven system. By systematically highlighting both research gaps and emergent topics (through BERTopic’s cluster terms), this approach is poised to improve global research efficiency and reinforce evidence-based medicine principles in numerous clinical and translational arenas.

\section{Limitations and Future Directions}
\label{sec:limitations}
Despite promising results, several constraints warrant further examination:
\begin{itemize}
    \item \textbf{Data Quality and Representativeness:} The accuracy of classification models and topic clustering heavily depends on the quality and diversity of training data. Underrepresented subdomains in BHI may see diminished performance.
    \item \textbf{Verification Overheads:} While continuous updates ensure currency, verifying the trustworthiness of new studies demands structured pipelines or human oversight. 
    \item \textbf{Scalability and Interoperability:} Large-scale real-world implementation may require robust interfaces with institutional databases, posing potential challenges around data privacy and standardization.
    \item \textbf{User Adoption Studies:} The system’s utility ultimately depends on acceptance by researchers and clinicians. Formal usability assessments and iterative improvements will be essential to refine the interface and outputs.
\end{itemize}

Future work will focus on more sophisticated machine learning workflows, stronger integration with hospital informatics systems, and user-centered design studies to optimize the platform’s real-world applicability.

\section{Conclusion}
\label{sec:conclusion}
Our AI-enhanced system reimagines systematic reviews in the Brain-Heart Interconnectome (BHI) by integrating advanced NLP methods, graph-based relationship modeling, and continuous database updates. Key functionalities—ranging from automated PICOS compliance detection for BHI to topic modeling—significantly reduce the burden of manual screening and refine research focus. By providing timely insights, democratized access via conversational AI, and robust data curation, the system delivers a transformative platform for evidence synthesis in BHI and potentially beyond. This technology-supported paradigm promises to uphold methodological rigor, mitigate research waste, and empower more decisive, up-to-date clinical and reporting guidelines' policy-making actions.

\bigskip


\end{document}